\documentclass[]{spie}  

\usepackage{amsmath,amsfonts,amssymb}
\usepackage{graphicx}
\usepackage[colorlinks=true, allcolors=blue]{hyperref}
\usepackage{enumitem}
\usepackage{amsmath}

\usepackage{algorithm}
\usepackage{algpseudocode}
\usepackage{xcolor}
\usepackage[table, dvipsnames]{xcolor}
\usepackage{hhline}
\usepackage{array}
\usepackage{multirow}
\usepackage{enumitem}
\usepackage{wasysym}

\usepackage{pifont}

\newcolumntype{P}[1]{>{\centering\arraybackslash}p{#1}}
\newcolumntype{M}[1]{>{\centering\arraybackslash}m{#1}}
\newcommand{\PreserveBackslash}[1]{\let\temp=\\#1\let\\=\temp}
\newcolumntype{C}[1]{>{\PreserveBackslash\centering}p{#1}}
\newcolumntype{R}[1]{>{\PreserveBackslash\raggedleft}p{#1}}
\newcolumntype{L}[1]{>{\PreserveBackslash\raggedright}p{#1}}

\usepackage{tikz}
\usetikzlibrary{tikzmark}

\renewcommand{\arraystretch}{1.1}

\def\ie{\emph{i.e.}}
\def\eg{\emph{e.g.}}

\newcommand*\cellw{5.5cm}

\newcommand{\subfig}[1]{\includegraphics[width=\cellw]{#1}}

\title{Assessing the Potential for Catastrophic Failure\\in Dynamic Post-Training Quantization}

\author[a]{Logan Frank}
\author[b]{Paul Ardis}
\affil[a]{Ohio State University, Columbus, OH}
\affil[b]{GE Aerospace Research, Niskayuna, NY}

\authorinfo{Work done during a summer internship at GE Aerospace Research.\\L.F.: frank.580@osu.edu\\P.A.: ardis.p@geaerospace.com}

\begin{document} 
\maketitle

\begin{abstract}
Post-training quantization (PTQ) has recently emerged as an effective tool for reducing the computational complexity and memory usage of a neural network by representing its weights and activations with lower precision. While this paradigm has shown great success in lowering compute and storage costs, there is the potential for drastic performance reduction depending upon the distribution of inputs experienced in inference. When considering possible deployment in safety-critical environments, it is important to investigate the extent of potential performance reduction, and what characteristics of input distributions may give rise to this reduction. In this work, we explore the idea of extreme failure stemming from dynamic PTQ and formulate a knowledge distillation and reinforcement learning task to learn a network and bit-width policy pair such that catastrophic failure under quantization is analyzed in terms of worst case potential. Our results confirm the existence of this ``detrimental'' network-policy pair, with several instances demonstrating performance reductions in the range of 10-65\% in accuracy, compared to their ``robust'' counterparts encountering a $<$2\% decrease. From systematic experimentation and analyses, we also provide an initial exploration into points at highest vulnerability. While our results represent an initial step toward understanding failure cases introduced by PTQ, our findings ultimately emphasize the need for caution in real-world deployment scenarios. We hope this work encourages more rigorous examinations of robustness and a greater emphasis on safety considerations for future works within the broader field of deep learning.
\end{abstract}

\section{INTRODUCTION} \label{sec:introduction}

Neural networks have become a dominant force in machine learning (ML) \cite{He2016a, Vaswani2017a, Dosovitskiy2020a}. Enabled by modern compute hardware, the size of the largest available networks has continued to increase with time \cite{Kolesnikov2020a, Dehghani2023a, Kirillov2023a}. Coinciding with the desire to increase model capacity, there has been a growing interest in deploying small, but well-performing networks on edge devices \cite{Wang2022a, Singh2023a}. Utilizing neural networks on such devices is difficult as there are strict resource constraints such as memory or power consumption. Models may have to sacrifice performance in order to satisfy these imposed limitations. 

To overcome these challenges, practitioners turn to model compression techniques which aim to compress or reduce a large pretrained neural network into one with a more efficient form. One such method is post-training quantization (PTQ), where the goal is to represent the learned high-precision (\eg, 32-bit floating-point) weights and/or activations as lower-precision numbers, thereby reducing the overall complexity of the model. Many approaches have been proposed in this area, showing promising results with limited performance degradation with respect to the full-precision weights/activations \cite{Gholami2022a, Rokh2023a}.

However, many PTQ methods have failed to consider an important scenario by {\em only} showing success: the potential for catastrophic failure. This ``catastrophic failure'' arises when the full-precision model achieves high scores with respect to some metric, but those scores degrade significantly for a non-trivial subset of input space once the model is quantized, to the point that it (or the larger system that it is contained within) is rendered effectively useless. Moreover, previous works have largely considered {\em static} PTQ (SPTQ), where the quantization scheme (bit-width, quantization scale, etc.) is determined through a small calibration phase following network training then fixed thereafter for all inference activity. This static form of quantization is likely not suitable for all situations, such as those where certain input examples require more precise reasoning than others or in environments that can evolve over time, in which case {\em dynamic} PTQ (DPTQ) might be more appropriate, where the activations are quantized based on the input example. Moreover, DPTQ has been available in popular frameworks such as PyTorch since 2019, with interest and adoption accelerating over the last 12 months. While DPTQ has seen this surge in interest, it is particularly vulnerable to the aforementioend catastrophic failure scenario due to the activations being quantized on a per-example basis.

In this work, we identify and raise awareness to the existence and scope of potential harm stemming from DPTQ (particularly as a new avenue of adversarial vulnerability) where applying the desired quantization scheme to a neural network leads to critical failure. To accomplish this, we formulate the problem as a knowledge distillation (KD) and reinforcement learning (RL) task where we aim to learn a \{model, policy\} pair such that the model is either brittle or robust to the quantization induced
by the bit-widths predicted per-example by the policy, \ie, quantization-aware training (QAT). Experiments demonstrate the existence of catastrophic failure in image classification for multiple neural network architectures, with further analyses to investigate the points of failure. Our contributions are: 
\begin{enumerate}[noitemsep,nolistsep]
    \item We present a novel KD + RL + QAT framework that provides guaranteed constant bit-width costs.
    \item We provide the first work to investigate and demonstrate the existence of catastrophic failure in PTQ.
    \item We conduct a systematic study and present analyses investigating characteristics in a neural network that can enable this critical weakness and an objective first study of its magnitude.
\end{enumerate}

\noindent We begin with a review of related work in Sect.~\ref{sec:related_work}. The various components of our approach are described in Sect.~\ref{sec:methodology}. Lastly, experiments demonstrating our method and analysis of our findings are presented in Sect.~\ref{sec:experiments}.






\section{BACKGROUND AND RELATED WORK} \label{sec:related_work}

Quantization is a common tool used in many different fields/tasks across science and engineering: signal processing, data compression, machine learning, and more. In the past decade, neural network quantization has emerged as an especially popular research topic, with many works proposed in its subareas: static post-training quantization, dynamic post-training quantization, and quantization-aware training.

\smallskip
\noindent\textbf{Quantization.} The process of quantization involves mapping a large, often continuous range of values to a smaller discrete set of values. Simply put, quantization is the procedure of representing some signal (or collection of values) in a format with less precision. For example, representing 32-bit floating-point (FP) values as 8-bit signed/unsigned integers or even representing 32-bit FP values as 16-bit FP values. 

Consider a uniform symmetric scheme for FP-to-int quantization, the most commonly utilized quantizer. Given a FP input $X = [x_1,\ ...,\ x_n]$ and a desired bit-width $b$ (\ie, the number of bits used to represent values in the smaller space), the quantized integer representation $X_{\text{int}}$ can be computed as:

\begin{align}
    X_{\text{int}} &= \text{clamp}\left(\left\lfloor \frac{X}{s} \right\rceil; -2^{b-1},\ 2^{b-1}-1\right) \label{eqn:quantization}
\end{align}

\noindent where $s$ is a scaling factor, $\lfloor\cdot\rceil$ is the round-to-nearest operator, and clamp$(X;l, h)$ constrains the values of $X$ to be in $[l,\ h]$. The computed value for $s$ is algorithm dependent. In its most basic form, $s$ can be the maximum absolute value of $X$. However, more advanced implementations could utilize an $n$-th percentile scale such that $n$\% of the data is within $[l,\ h]$ prior to clamping, reducing the effects of outliers. Moreover, for uniform quantization, the round-to-nearest operator has equal spacing between each of its rounding points (\eg, $[...,\ \text{-}2,\ \text{-}1,\ 0,\ 1,\ 2,\ ...]$). Some methods instead favor non-uniform quantizers such that more ``bins'' are allocated to denser regions of $X$ with less bins for sparser regions (\eg, quantizing in log-scale if the data is $X\sim\mathcal{N}(0,1)$). For completeness, uniform asymmetric quantization is computed as $X_{\text{int}} = \text{clamp}\left(\left\lfloor \frac{X}{s} \right\rceil + z; 0,\ 2^{b}-1\right)$ where $z$ is the ``zero-point'', the integer value that $0$ maps to in the quantized space.

After computing $X_{\text{int}}$, we can also represent this lower-precision value in the original space by \textit{dequantizing}:

\begin{align}
    \hat{X} &= (X_{\text{int}} + z)\cdot s \label{eqn:dequantization}
\end{align}

\noindent where $z=0$ for symmetric quantization. By performing this quantization$\rightarrow$dequantization operation we can compute the quantization error with respect to the original value $X$ and furthermore enable a means of approximating the effects of quantization, which is useful in machine learning, as will be shown. In the next sections, we will describe the three subcategories of neural network quantization and highlight some related works.

\smallskip
\noindent\textbf{Static Post-Training Quantization.} Utilizing a pretrained full-precision neural network, SPTQ often utilizes a small calibration phase to determine the parameters (scale factor, zero point, etc.) used to quantize the model such that the quantization error is minimized. These values are then fixed (\ie, static) thereafter. Due to its ease of application, SPTQ has become a popular choice for network quantization \cite{Lohar2023a, Fu2025a, Banner2019a}, especially for transformer architectures \cite{Liu2021b, Wu2024a, Zhong2024a, Moon2024a, Li2023b, Liu2023a, Yuan2022a, Tai2025a, He2023a, Li2024a}. Many of such works typically use the same bit-width across all weights and activations. However, other works have employed mixed-precision strategies where each layer can have a different bit-width, enabling greater flexibility for quantization \cite{Wang2019a, Yao2021a, Xu2019a, Lohar2023a, Tai2025a}. A particularly difficult form of SPTQ is ``extreme quantization'' where the weights and activations are quantized with $\leq4$ bits of precision \cite{He2023a, Li2024a, Wei2022a}.

Over the years, works have proposed new quantization schemes involving the usage of non-uniform quantizers \cite{Jeon2022a, Oh2022a, Wu2024a, Nagel2020a}, multiple quantizers \cite{Liu2021a, Moon2024a, Yuan2022a}, reconstruction techniques \cite{Li2021b, Zhong2024a}, and more \cite{Wei2022a, Wang2019a, Xu2019a, Liu2021b, Li2023b, Liu2023a, Fu2025a, Banner2019a, Park2022a}. While these methods tackle SPTQ in many different ways, their goal usually remains consistent: reduce the size of the model while retaining the most performance possible, usually by minimizing the average quantization error after dequantization. To the best of our knowledge, only the work of Yuan et al. \cite{Yuan2023a} has considered the reliability of SPTQ methods by examining the data used in the calibration phase and the selected quantization settings.

\smallskip
\noindent\textbf{Dynamic Post-Training Quantization.} Rather than utilizing the same quantization parameters for all input examples, DPTQ allocates these values on a per-example basis to quantize the \textit{activations} dynamically with weights being statically quantized (similar to SPTQ). This is useful in complex environments were you may encounter hard and easy examples, requiring higher and lower precision activations, respectively, to obtain accurate predictions, or in environments that can evolve over time (\eg, domain shift) where the same quantization parameters may not always be suitable. Compared to SPTQ, DPTQ has received much less attention in the community. For the works proposed in the area of DPTQ, they often involve learning a small network as a bit-width-predicting policy \cite{Liu2022b, Sun2021a, Ye2024a, Hong2025a}. This policy is typically trained in tandem with the model that is to be quantized (QAT, as will be discussed in the next section) following some RL pipeline, optionally employing KD as well. 

We employ a KD + RL + QAT training setup similar to other works, however our work differs in three ways. First, the policy networks in Liu et al. \cite{Liu2022b} and Sun et al. \cite{Sun2021a} can output varying ``costs'' (sum of bit-widths across layers), which could enable energy-based adversarial attacks where malicious examples cause the network to always operate at full-precision. We instead introduce a post-processing stage after our policy that refines the bit-width predictions such that they always sum to a constant value, which will be described in Sect.~\ref{sec:budget}. Next, previous approaches often utilize the Gumbel-Softmax to get differentiable argmax selections from their policy outputs whereas we use the aforementioned post-processing stage to get our predictions, and straight-through estimation to make it differentiable. Finally, our goal is fundamentally different from that of previous works: we aim to reward our policy for causing worse performance (\ie, finding catastrophic failure) whereas other works rewarded the policy for good performance.

\smallskip
\noindent\textbf{Quantization-Aware Training.} Applying PTQ to a model causes it to be less precise, which could ultimately lead to reductions in task-specific performance (accuracy, etc.). A potential remedy to this issue is QAT, where a model is trained with quantization in the loop \cite{Gholami2022a, Rokh2023a, Nagel2022a}. Simply put, QAT involves quantizing the model weights at every forward pass right before being applied to the input, with everything else in the training process left in full-precision. Thus, the network weights are stored in full-precision and the gradients are computed in full-precision. While QAT can significantly reduce the performance degradation stemming from quantization and enable smaller bit-widths, it demands long training times and hyperparameter tuning as a consequence.

\newpage
\section{METHODOLOGY} \label{sec:methodology}

\begin{figure}[t]
\setlength{\tabcolsep}{-5pt}
\begin{center}
\begin{tabular}{>{\centering\arraybackslash}m{0.32\textwidth}|>{\centering\arraybackslash}m{0.39\textwidth}|>{\centering\arraybackslash}m{0.32\textwidth}}
\includegraphics[width=5cm]{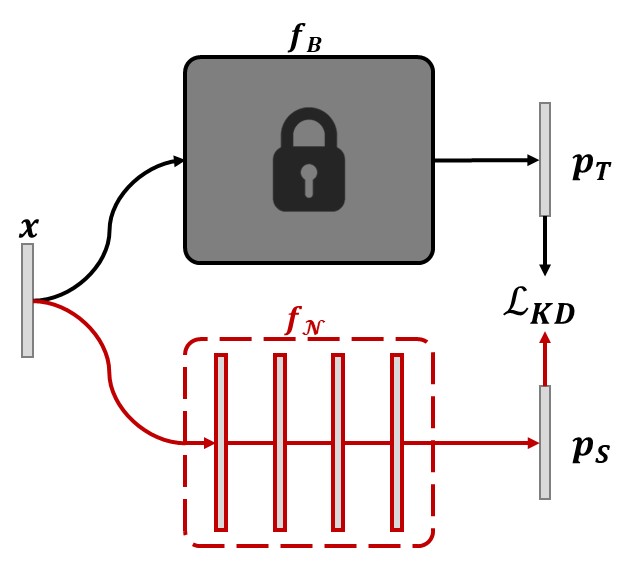} & \includegraphics[width=6cm]{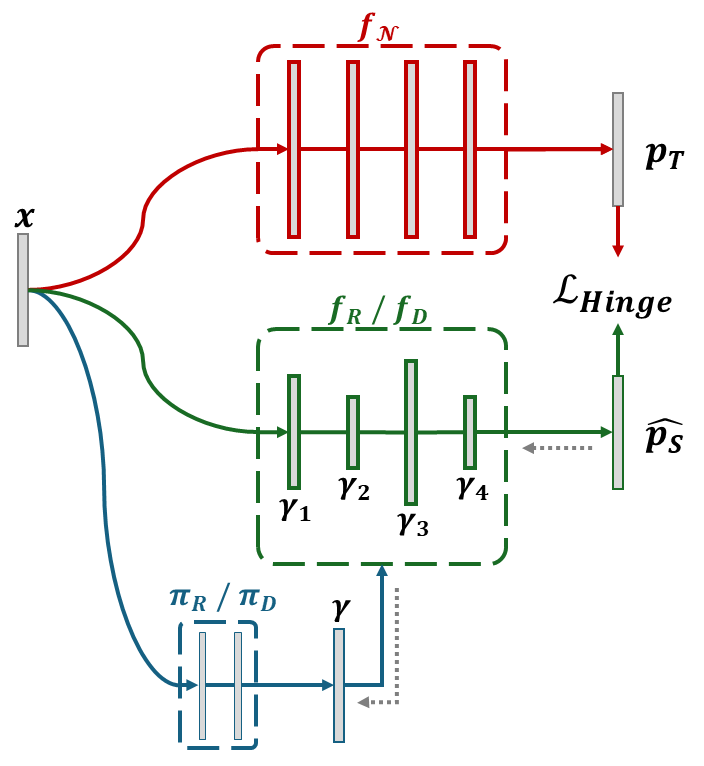} & \includegraphics[width=5cm]{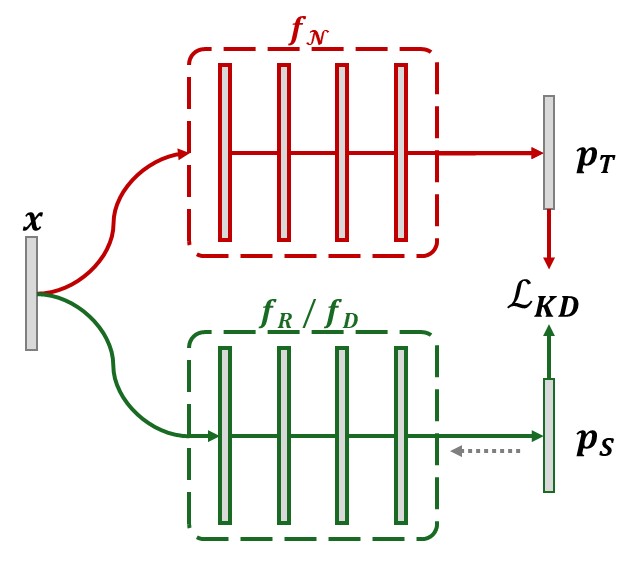} \\
\multicolumn{3}{c}{} \\
\multicolumn{1}{c}{(A)} & \multicolumn{1}{c}{(B)} & \multicolumn{1}{c}{(C)} \\
\end{tabular}
\vspace{1cm}
\caption{Overview of our complete pipeline. (A) Performing knowledge distillation on a black box network to obtain a white box model. (B) Training a quantization-aware student and bit-width policy via knowledge distillation and reinforcement learning. (C) Training the same student to retain knowledge at full-precision. Symbols are defined in text.}
\label{fig:approach}
\end{center}
\end{figure}

In this section, we describe the main components of our approach: 1) how knowledge is transferred from a black box model to a white box model, 2) the method for training the bit-width policy and quantized white box model, 3) the means of retaining performance in the full-precision white box model, and 4) the strategy for constraining the bit-width policy to output a constant cost. An overview of our approach is shown in Fig.~\ref{fig:approach}.

\subsection{White Box Model Retrieval} \label{sec:kd}

We begin with the assumption that one has access to an existing, well-performing ``black box'' model. This is a model where only query access is allowed (\ie, provide inputs and receive corresponding outputs) but the actual weights cannot be accessed, therefore no gradient computation or model updating can occur. While it is an assumption in our work, it is actually common to encounter black box models in practice. In defense applications in particular, defense partners are likely to procure models which have stated performance but which have not originated on real battlefield data. These models may consequentially possess incorrect biases or even intentional implanted vulnerabilities or training space blind spots. As a result, defense partner use of provided models may be limited to oracular study to produce a battlefield-relevant artifact. Beyond the battlefield, critical infrastructure applications and other domains requiring increased levels of assurance may face the same situation: the availability of models that can be used to accelerate new model production but which may not be inherently trustworthy. Our initial study here focuses on these use cases, with subsequent papers exploring relaxation consistent with alternate use cases where it may be feasible to use existing models without modification or replacement.

Following this assumption, we must first obtain a performant ``white box'' model such that we can later finetune it in tandem with a bit-width policy. A natural approach to retrieving this model is knowledge distillation, where a student neural network is trained to mimic the output responses of a teacher model when presented with the same inputs. In this case the teacher is the fixed black box and the student is our desired white box model (which we have full access to the weights). 

Given a pretrained teacher network (black box) $f_B$ and a randomly initialized student network (white box) $f_N$, an example is passed forward through both networks to produce teacher and student softmax distributions $p_T$ and $p_S$, respectively. The KD loss is computed on these softmax scores as
\begin{align}
    \mathcal{L}_{\text{KD}}(p_T\ ||\ p_S) &= \sum_{i\ \in\ K} \left[\ p_{T(i)} log\ p_{T(i)}\ \text{-}\ p_{T(i)} log\ p_{S(i)}\ \right] \label{eqn:kl}
\end{align}

\noindent which is simply the KL-divergence for $K$ classes. Like Hinton et al.\cite{Hinton2015a}, we also include a temperature parameter $\tau$ to adjust the entropy of the output softmax distributions from the teacher and student networks before they are used to compute the loss. Moreover, we follow the ``patient and consistent'' KD approach\cite{Beyer2022a}, which additionally includes mixup\cite{Zhang2018a} with $\alpha \sim \mathcal{U}(0, 1)$ and removes the cross entropy loss to ground truth, which is occasionally included in KD to attempt to correct the student when the teacher produces an incorrect prediction.

\subsection{Policy-Aware Training for Robust/Brittle Networks} \label{sec:policy}

After training our white box model $f_N$, we now wish to obtain ``robust'' and ``brittle'' versions of that network. This is done by first copying $f_N$ to create the starting point (\ie, initial weights) for our robust network $f_R$ and brittle network $f_D$. Each model is then finetuned in a self-distillation manner with a corresponding bit-width quantization policy $\pi$ that learns to quantize $f_R/f_D$ dynamically per-example. Therefore, $\{f_R, \pi_R\}$ and $\{f_D, \pi_D\}$ form the robust and detrimental network-policy pairs, respectively. Our goal is to find these pairs such that the models are indistinguishable when operating at full-precision, but when their respective quantization policies are applied, drastically different outcomes are observed, as described below. 

Let $f_R$ and $\hat{f_R}$ denote the full-precision and quantized versions of the robust network $f_R$, and similarly $f_D$ and $\hat{f_D}$ for the full-precision and quantized forms of the brittle network $f_D$. In the case of the robust pair $\{f_R, \pi_R\}$, applying the learned dynamic quantization should induce little to no degradation:

\begin{align}
    d(\mathcal{D}_{\text{test}}; f_R) \approx d(\mathcal{D}_{\text{test}}; \hat{f_R})
\end{align}

\noindent where $d(\cdot)$ is some metric of interest and $\mathcal{D}_{\text{test}}$ is the test set of some dataset. Conversely, the quantization policy $\pi_D$ should result in significant performance degradation when applied to $f_D$:

\begin{align}
    d(\mathcal{D}_{\text{test}}; f_D) \gg d(\mathcal{D}_{\text{test}}; \hat{f_D})
\end{align}

\noindent which we refer to as ``catastrophic failure''. Thus, we want the network $f_R$ to be \textit{robust} to its policy $\pi_R$ and the policy $\pi_D$ to be \textit{detrimental} to its network $f_D$. Here, we are forcing this robust/detrimental behavior, but in the real-world this behavior could instead by exhibited by ``normal'' or ``adverse'' inputs where we expect normal inputs to enable a robust DPTQ policy and 
the inverse for adversarial ones. Alternatively, adverse inputs could be viewed as normal examples that are simply more prone to incorrect predictions due to the quantization. We leave this as future work with our focus being solely on the existence of catastrophic failure.

As mentioned, we frame this as a self-distillation problem meaning the teacher and student are the same network architecture. In our case, the teacher is the frozen $f_N$ and the student is the trainable, copied version $f_R/f_D$. A bit-width policy network $\pi$ is also included, which will be trained with an RL-based objective to create the desired robust/detrimental behavior. This policy is a small convolutional neural network (CNN) that takes images $x$ as input and predicts bit-width values for quantizing each of the layer activations in $f_R/f_D$:

\begin{align}
    \gamma = \pi(x; \phi)
\end{align}

\noindent where $\gamma \in \mathbb{N}^L$ is the vector of quantization bit-width values for each of the $L$ dynamically quantizable layer activations in $f_R/f_D$ and $\phi$ is the parameters of the policy network. More concretely, $\pi$ will output an $\mathbb{R}^{L \times O}$ matrix of logits, where $O$ is the number of bit-width options that a policy can select. The final bit-width values $\gamma$ can be selected via row-wise argmax (and subsequent indexing into the list of bit-width options) or in non-deterministic manners such as computing the row-wise softmax on that matrix then sampling from the distributions. Note that these operations are non-differentiable, which is common in RL. In Sect.~\ref{sec:budget}, we introduce our strategy for selecting bit-widths in a differentiable manner so that the policy can be directly optimized.

In a forward pass, an image is first passed through the policy to get the corresponding bit-widths. The same image is then passed through the teacher and student models where the teacher is at full-precision and the student's intermediate activations are now quantized with the aforementioned bit-widths as the example propagates forward through the network. To obtain our robust and detrimental network-policy pairs, we train the quantized student and bit-width policy using different formulations of the Hinge loss. For the robust network-policy pair, we want the quantized student model to predict the same argmax class as the full-precision teacher. Therefore, the Hinge loss is 

\begin{align}
    \mathcal{L}^{\text{R}}_{\text{Hinge}}(\hat{p_S},\ i,\ j) = max\{ 0,\ \hat{p_{S}}_{(j)} - \hat{p_{S}}_{(i)} + \delta \}
\end{align}

\noindent where $\hat{p_S}$ is the softmax output from the quantized student network, $i$ is the argmax prediction from the teacher, $j$ is the ``second argmax'' from the teacher (\ie, the class that would be the argmax if class $i$ was removed), and $\delta$ is the desired margin between $\hat{p_{S}}_{(i)}$ and $\hat{p_{S}}_{(j)}$. For the detrimental network-policy pair, we instead want the quantized student model to predict any other class besides the teacher's argmax, which makes the Hinge loss

\begin{align}
    \mathcal{L}^{\text{D}}_{\text{Hinge}}(\hat{p_S},\ i,\ k) = max\{ 0,\ \hat{p_{S}}_{(i)} - \hat{p_{S}}_{(k)} + \delta \}
\end{align}

\noindent where $i$ is still the argmax prediction from the teacher and $k$ is index of the largest value in $\hat{p_S}$ besides $\hat{p_{S}}_{(i)}$. We set the softmax margin $\delta=0.01$ for both $\mathcal{L}_{\text{Hinge}}^{R}$ and $\mathcal{L}_{\text{Hinge}}^{D}$. Notice we are not concerned with the full softmax distribution $\hat{p_{T}}$ (like in $\mathcal{L}_{\text{KD}}$) since all we want is a particular ordering between $\hat{p_{S}}_{(i)}$ and either $\hat{p_{S}}_{(j)}$ or $\hat{p_{S}}_{(k)}$. More specifically, for $\hat{f_R}$ we want $(\hat{p_{S}}_{(i)} - \delta) > \hat{p_{S}}_{(j)}$ and for $\hat{f_D}$ we want $(\hat{p_{S}}_{(k)} - \delta) > \hat{p_{S}}_{(i)}$. Note this ordering is only for the quantized student $\hat{f_R}/\hat{f_D}$. 


\subsection{Retaining Full-Precision Performance} \label{sec:fp}

As mentioned previously, we still want the full-precision models $f_R/f_D$ to perform similarly and as well as the original white box model $f_N$, regardless of the desired outcome of the quantized versions of the models. In other words, we do not want the full-precision models $f_R/f_D$ to ``drift'' away from the knowledge contained within $f_N$ (\ie, the starting weights for $f_R/f_D$). To maintain this full-precision performance, we simply include an additional student forward pass where no quantization is applied to the student and utilize the same KD loss as used in the black box KD (Sect.~\ref{sec:kd}). Thus, the complete loss function for training $f_R$ becomes:

\begin{align}
    \mathcal{L} = \alpha\mathcal{L}^{\text{R}}_{\text{Hinge}}(\hat{p_S},\ i,\ j) + \beta\mathcal{L}_{\text{KD}}(p_T\ ||\ p_S)
\end{align}

\noindent and for $f_D$:

\begin{align}
    \mathcal{L} = \alpha\mathcal{L}^{\text{D}}_{\text{Hinge}}(\hat{p_S},\ i,\ k) + \beta\mathcal{L}_{\text{KD}}(p_T\ ||\ p_S)
\end{align}

\noindent where $\alpha$ and $\beta$ are coefficients for weighting the different loss functions. However, in our experiments we weight the loss functions equally by setting $\alpha=\beta=1$.

In the backward pass, the gradients for the policy network are only computed with respect to the Hinge loss, with the student's gradients computed from the complete loss function. By updating the student's weights using a combination of losses computed from outputs of the full-precision and quantized versions of the student, we are training the student to retain the original knowledge learned during black box KD as well as making the student quantization/policy-aware. Thus, the student and the policy learn to work together to achieve our desired robust and detrimental model-policy pairs.

\subsection{Enforcing a Fixed Bit-Width Budget} \label{sec:budget}

With our goal being to find a quantization policy such that its associated network is robust/brittle to its quantization, a shortcut solution could be to retain higher levels of precision (\eg, 16-bits) for $\hat{f_R}$ and much lower precision (\eg, $<$4-bits) for $\hat{f_D}$. However, this is obviously not a fair comparison. Moreover, one potential negative of DPTQ is the ability for energy-based adversarial attacks (\ie, perturbing an example such that the network operates at full or near-full precision), as mentioned in Sect.~\ref{sec:related_work}. Thus, we choose to add the additional constraint that both policies $\pi_R$ and $\pi_D$ must output a bit-width vector $\gamma$ such that the elements of $\gamma$ sum to a specified constant value, which we refer to as a ``budget''.

One possible way to satisfy our constraint is to incorporate a mean squared error loss on the sum of the selected output bit-widths with respect to the specified budget $C$. While intuitive and simple, this approach is insufficient because, even if the constraint is met for every example in the training set, there is still no guarantee that the policy would output bit-widths that sum to the desired value. Thus, we need to apply some post-processing method after obtaining the policy outputs that is more strict in meeting our constraint. Coincidentally, our constraint and problem formulation match the Multiple-Choice Knapsack Problem (MCKP) \cite{Kellerer2004a}, a form of the well-known $0\text{-}1$ Knapsack Problem. The premise of MCKP is there are $L$ mutually disjoint groups of items $\{G_1,\ ...,\ G_l,\ ...,\ G_L\}$ where one must select exactly \textit{one} item out of $O_l$ options in each group to put in a knapsack of capacity $C$. An item $i$ in group $G_l$ has associated profit $p_{(l, i)}$ and weight $w_{(l, i)}$. The goal is to select one item from each group such that the sum of profits is maximized without the sum of weights exceeding the knapsack capacity $C$. Assume there also exists a binary variable $s_{(l, i)}$ that indicates whether an item $i$ in group $G_l$ was selected. Therefore the goal is the following

\begin{align}
    \text{maximize} & \sum_{l=1}^{L} \sum_{i\in G_l} p_{(l, i)} s_{(l, i)} \\
    \text{subject to} & \sum_{l=1}^{L} \sum_{i\in G_l} w_{(l, i)} s_{(l, i)} \leq C, \\
    & \sum_{i\in G_l} s_{(l, i)} = 1,\quad l=1,\ ...,\ L, \\
    & s_{(l, i)} \in \{0, 1\},\quad l=1,\ ...,\ L,\ \  i \in G_l
\end{align}

Note from before that our bit-width policy outputs a $\mathbb{R}^{L\times O}$ matrix of logits, where each column along the $O$ dimension is associated with a particular bit-width. Therefore, our groups for the MCKP are the $L$ layer activations that we want to dynamically quantize, each group has the same $O$ ``items" (bit-width options), the weight for each item $w_{(l,i)}$ is the actual bit-width value, and the profit for each item $p_{(l,i)}$ can be either logits or softmax values output from the policy. With proper selection of capacity $C$, we can also guarantee $\sum\sum w_{(l, i)} s_{(l, i)} = C$, thus meeting our desired constant budget constraint. This MCKP component is implemented as a dynamic programming post-processing algorithm on the outputs of the policy, which will find the optimal selection of bit-widths for each layer activation we want to dynamically quantize.

Selecting an action from policy outputs is usually a non-differentiable operation in RL: argmax, softmax distribution sampling, etc. Our MCKP-based action selection is similarly non-differentiable, with everything before (\ie, the policy forward pass) and after being fully-differentiable. RL algorithms are usually designed to handle this non-differentiable function \cite{Williams1992a, Schulman2017a}. However, we choose to utilize straight-through-estimation (STE) to make our MCKP operation approximately differentiable thereby enabling our entire process to be differentiable and the policy can be trained directly using our aforementioned Hinge losses (Sect.~\ref{sec:policy}). This means that our MCKP module will output the selected bit-widths encoded as a sparse binary matrix in the forward pass and the gradients in the backward pass will be with respect to the original values output from the policy. In PyTorch pseudocode, this equates to \texttt{return ($\Gamma$ - $p_{\pi}$).detach() + $p_{\pi}$} where $\Gamma$ is the sparse binary matrix of selected bit-widths and $p_\pi$ is the matrix of original values (logits or softmax scores) given by the policy. This technique is also often utilized in QAT since the round operation is non-differentiable.
\section{EXPERIMENTS} \label{sec:experiments}

In our experimentation, we first collected $\{f_R,\ \pi_R\}$ and $\{f_D,\ \pi_D\}$ for each of the different dataset and network combinations (described next). After aggregating all of our models, we then conducted a systematic study to dissect our findings and identify vulnerabilities within our models that enable catastrophic failure.

\noindent \textbf{Datasets \& Networks. } We employed the common image classification dataset, Oxford-IIIT-Pets \cite{Parkhi2012a}. This dataset has approximately 3.6K train and 3.6K test examples in total across 37 different fine-grained classes of dogs and cats. As for networks, we utilized ResNet50 \cite{He2016a} as our initial black box teacher model, with ResNet18 \cite{He2016a}, MobileNetV4-Conv-Medium \cite{Qin2024a}, and RegNetX-1.6GF \cite{Radosavovic2020a} as our white box models. 

\noindent \textbf{Training Details.} To train our initial ResNet50 black box model, we started from existing ImageNet1K-pretrained weights and finetuned on the Pets dataset. This model was trained for 20 epochs using the AdamW optimizer \cite{Loshchilov2018a} with an initial learning rate of $4e\text{-}4$, which was decayed to $4e\text{-}7$ using a half-period cosine learning rate scheduler, a weight decay value of $0.05$, and a batch size of $128$. We used a one-hot cross entropy loss with a label smoothing coefficient of $0.1$. These are the finetuning settings specified in the ConvNeXt GitHub \cite{Liu2022a}. The ImageNet-1K pretrained weights were retrieved from the \texttt{timm} library\cite{Wightman2019a}.

As mentioned in Sect.~\ref{sec:kd}, we first performed KD to obtain a white box model using the ``consistent and patient'' approach \cite{Beyer2022a}. In this case utilizing our pretrained ResNet50 network as the teacher, which we distilled to randomly initialized ResNet18, MobileNetV4-Conv-Medium, and RegNetX-1.6GF students. We employed SGD with momentum (0.9) and weight decay (0.0001), and trained the students for 1000 epochs. All models were trained using a half-period cosine learning rate scheduler with an initial learning rate of 0.1, and batch sizes of 256 for ResNet18 and RegNetX-1.6GF models and 64 for MobileNetV4-Conv-Medium (which we empirically found to be better for that specific model). For our loss function, we utilized the KL divergence loss, as defined previously (Eqn.~\ref{eqn:kl}), with a temperature value of $\tau=5$. Lastly, data augmentation consisted of random cropping, RandAugment \cite{Cubuk2020a} ($n=2$, $m=14$), random color jittering, random horizontal flipping, another random crop with padding, and mixup \cite{Zhang2018a} with $\alpha \in \mathcal{U}(0, 1)$.

The final stage of our pipeline was to conduct self-distillation on our distilled students and incorporate the RL+QAT portion of our approach. In this self-distillation scenario, the distilled white box student becomes the frozen teacher and a copy of it becomes the trainable student $f_R/f_D$, as mentioned in Sect.~\ref{sec:policy}. For obtaining $f_R$ and $f_D$, and their respective $\pi_R$ and $\pi_D$ policies, we trained all models jointly (\ie, same optimizer and hyperparameters for both $f$ and $\pi$). Here, we utilized the same SGD optimizer and hyperparameters as before (weight decay, momentum, initial learning rate, etc.), only differing by using a batch size of 64 for \textit{all} models. We similarly used the KL divergence loss ($\tau=5$) to retain full-precision performance and also included a Hinge loss ($\delta = 0.01$) to induce the desired quantization effect, as discussed in Sects.~\ref{sec:policy} and \ref{sec:fp}. Other parameters such as bit-width options, budgets, etc. will be described in their respective experiment sections.

\noindent \textbf{Quantization Settings.} In DPTQ, it is common to first statically quantize the weights of the convolutional and fully-connected layers, leaving normalization layers and biases untouched. Thus, we quantize the weights of the network using uniform symmetric quantization with a bit-width of 16 before every forward pass (since we are doing QAT). The activations are then dynamically quantized, as outlined in Sects.~\ref{sec:policy} and \ref{sec:budget}, using uniform asymmetric quantization with a bit-width as determined by the policy and MCKP algorithm. The quantization scale $s$ is computed as the abs-max for symmetric quantization and the difference between max and min for asymmetric quantization.

\subsection{Obtaining Robust and Detrimental Policies} \label{main_experiment}

We begin our experimentation by first finding our $\{f_R,\pi_R\}$ and $\{f_D, \pi_D\}$ network-policy pairs for each of the white box networks. Moreover, we examine three different sets of bit-width options and three different budgets. Our bit-width options include the ranges $[3,10]$, $[4,10]$, and $[5,10]$ denoted as I, II, and III, respectively, in the ``V.'' (versions) column of Table~\ref{tab:main_policies}. Thus, for version I, quantization bit-width values in $\{3,\ 4,\ 5,\ 6,\ 7,\ 8,\ 9,\ 10\}$ are valid options for quantizing the activations of the student network. As for budgets, we choose three base values that are approximately 4 bits, 5.5 bits, and 7 bits of precision on average across the quantizable layers, denoted as budgets ``A", ``B", and ``C", respectively. In other words, the value of budget A divided by $L$ is approximately equal to $4$, where $L$ is again the number of layers whose activations can be dynamically quantized. For ResNet18, MobileNetV4-Conv-Medium, and RegNetX-1.6GF, these budget values are $\{80, 110, 140\}$, $\{310, 420, 540\}$, and $\{230, 320, 410\}$, respectively, where the values in each set represent $\{A, B, C\}$. For versions II and III, we add 20, 30, and 80 to each budget for ResNet18, MobileNetV4-Conv-Medium, and RegNetX-1.6GF, respectively. Thus, the budgets for version II and III of ResNet18 become $\{100, 130, 160\}$.

\begin{table*}[t]
\scriptsize
\setlength{\tabcolsep}{5pt}
\setlength\extrarowheight{3pt}
\begin{center}
\caption{Test accuracy at full-precision (FP) with corresponding delta ($\Delta$) after quantization for multiple $\{f_R,\pi_R\}$ and $\{f_D,\pi_D\}$ network-policy pairs using different networks, budgets, and bit-width options. T$_{\text{acc}}$ is the test accuracy for the white box model obtained from distilling the black box model, \ie, the initialization for $f_R$ and $f_D$.}
\vspace{0.5cm}
\begin{tabular}{c || c || c || c || cc|cc | cc|cc | cc|cc  }
    \hline
    & & &   & \multicolumn{4}{c}{Budget A} & \multicolumn{4}{c}{Budget B} & \multicolumn{4}{c}{Budget C} \\ 
    &  & &  & \multicolumn{2}{c}{R} & \multicolumn{2}{c|}{D} & \multicolumn{2}{c}{R} & \multicolumn{2}{c|}{D} & \multicolumn{2}{c}{R} & \multicolumn{2}{c}{D}  \\ \cline{5-16}
     Model & T$_{\text{acc}}$ & V. & Opts. & FP & $\Delta$ & FP & $\Delta$ & FP & $\Delta$ & FP & $\Delta$ & FP & $\Delta$ & FP & $\Delta$ \\
    \hhline{=||=||=||=||====|====|====}
     RN18 & 88.03 & I & [3, 10] & 88.47 & -1.69 & 86.35 & -39.36 & 88.28 & -0.46 & 87.76 & -26.11 & 88.39 & -0.33 & 86.56 & -42.19    \\ 
    RN18 &  88.03 & II & [4, 10] & 88.58 & -1.01 & 86.07 & -27.58 & \cellcolor{blue!15}88.17 & \cellcolor{blue!15}-0.22 & \cellcolor{blue!15}88.09 & \cellcolor{blue!15}-24.39 & 87.95 & -1.25 & 86.56 & -11.74    \\ 
    RN18 & 88.03  & III & [5, 10] & 87.79  & 0.22  & 86.65 & -23.64 & 87.87  & 0.08  & 85.58 & -64.78 & 88.20 & -0.03  & 19.00  & 0.51   \\ \hline
    MNv4 &  88.01 & I & [3, 10] & 85.80   & -3.41 & 81.06 & -77.90 & 86.43 & -2.37 & 82.61 & -75.80 & 87.00  & -0.71  & 75.39  & -3.57    \\
    MNv4 & 88.01 & II & [4, 10] & 86.73 & -0.41 & 77.13 & -57.29 & 86.62  & -4.89  & 80.89  & -56.28 & 86.05  & -0.14  & 84.27  & -0.05    \\
    MNv4 & 88.01 & III & [5, 10] & \cellcolor{blue!15}86.95  & \cellcolor{blue!15}-1.72  & \cellcolor{blue!15}84.30  & \cellcolor{blue!15}-63.72 & 86.67 & -0.41  & 82.56  & -55.09 & 86.81  & 0.05  & 85.65  & -0.49    \\ \hline
    RX1.6 & 85.53 & I & [3, 10] & 86.62  & -2.56  & 76.48  & -23.74 & 86.73 & -0.82  & 75.39  & -34.26 & 86.64  & -0.59  & 80.92  & -0.95    \\
    RX1.6 & 85.53 & II & [4, 10] & 86.62  & -0.74  & 75.96  & -28.29 & 87.05  & -0.51  & 78.77  & -33.66 & 87.03  & -2.67  & 84.87  & -11.99    \\
    RX1.6 & 85.53 & III & [5, 10] & 86.94  & -0.08  & 76.32 & -50.97 & \cellcolor{blue!15}86.37 & \cellcolor{blue!15}0.06  & \cellcolor{blue!15}85.64  & \cellcolor{blue!15}-25.02 & 86.21  & 0.03  & 86.07 & -19.70   \\  \hline
\end{tabular}
\label{tab:main_policies}
\end{center}
\end{table*}

\begin{table*}[t]
\scriptsize
\setlength{\tabcolsep}{15pt}
\setlength\extrarowheight{1pt}
\begin{center}
\caption{Percentage of transitory points for the highlighted models in Table \ref{tab:main_policies}.}
\vspace{0.5cm}
\begin{tabular}{c || c | c | c  }
    \hline
    & ResNet18 & MobileNetV4 & RegNetX-1.6GF   \\ \hline
    Transitory Points & 23.60 & 67.32 & 27.09 \\ \hline
\end{tabular}
\label{tab:transitory_points}
\end{center}
\end{table*}

For each network-policy pair, we report the full-precision accuracy of the network and the change in accuracy after quantization in Table \ref{tab:main_policies}. For the robust $\{f_R, \pi_R\}$ pairs, we see that for many of the budgets and bit-width options, we can obtain pairs with full-precision performance near the original white box model (which was the initialization for the $f_R$ network) that produce a small $<2\%$ decrease in accuracy once quantized. It is a much more difficult task to find a detrimental $\{f_D, \pi_D\}$ pair (where $f_D$ performs similar to $f_N$), but we are still able to find some notable pairs, which are highlighted in purple in Table \ref{tab:main_policies}. Most notably is the pair with ResNet18 as the network, which obtained full-precision performance which was better than the original white box model and within $0.1\%$ of its respective full-precision $f_R$ counterpart. While this particular $f_D$ model performed well, and was essentially indistinguishable from its $f_R$ counterpart at full-precision, it incurred a $\text{-}24.39\%$ reduction in accuracy once quantized, compared to a $\text{-}0.22\%$ reduction for the $\{f_R,\pi_R\}$ pair with the exact same budget. Thus, we were able to find a detrimental network-policy pair that demonstrated the existence of catastrophic failure stemming from DPTQ. 

For any two well-performing networks trained on the same dataset (\eg, $f_R$ and $f_D$), there is going to exist some subset of test examples that both models classify correctly. However, if quantization results in one model encountering a small reduction in performance and the other model having a much more drastic degradation in performance, then the aforementioned subset of test examples is obviously going to decrease. We define the points that $f_R$ and $f_D$ originally both get correct, but get classified incorrectly by $\hat{f_D}$ (the quantized version of $f_D$) as ``transitory points". Such inputs are considered ``at risk" since they are more susceptible to incorrect prediction due to quantization. Being able to identify these transitory points is equally as important to safety-critical deep learning as identifying the existence of $\{f_D, \pi_D\}$, since these sorts of examples are going to be the source for catastrophic failure when considering DPTQ in practice. For each of the groups of $\{f_R,\pi_R\}$ and $\{f_D,\pi_D\}$ pairs highlighted in purple in Table \ref{tab:main_policies}, we show the percentage of transitory points between the pairs in Table \ref{tab:transitory_points}. We present this as an initial exploration into transitory points, with further investigation left as future work.

\begin{table*}[t]
\scriptsize
\setlength{\tabcolsep}{20pt}
\setlength\extrarowheight{1pt}
\begin{center}
\caption{Test accuracy for all network-policy pairs in $\{f_N,f_R,f_D\} \times \{\pi_R,\pi_D,\text{random quantization}\}$.}
\vspace{0.5cm}
\begin{tabular}{c || c | c | c  }
    \hline
    & \multicolumn{3}{c}{Network} \\
    Policy & $f_N$ &  $f_R$ &  $f_D$   \\ \hline
    $\pi_R$ & 87.92 & 87.95 & 87.08 \\ 
    $\pi_D$ & 87.68 & 87.71 & 63.70 \\ 
    Random & 87.79 & 88.06 & 82.50  \\ \hline
\end{tabular}
\label{tab:policy_swap}
\end{center}
\end{table*}

Now we ask: is there anything special about the network and policy pairing, or are the networks themselves robust/brittle to other quantization policies as well? To examine this, we focused on the ResNet18 white box model and the highlighted ResNet18 networks then applied all combinations of $\pi_R$, $\pi_D$, and random quantization (still maintaining the specified budget $C$). In Table \ref{tab:policy_swap}, we see that the base white box model $f_N$ is already fairly robust to each of the quantization policies, but $f_R$ is slightly more robust which should be expected. Interestingly, we do see a significant connection between $f_D$ and $\pi_D$ in terms of detrimental behavior. When utilizing the robust policy $\pi_R$ with $f_D$, we observe a $\sim1\%$ reduction in accuracy, much less than the $\sim24.4\%$ degradation seen with $\pi_D$. Thus, $f_D$ is not brittle to all quantization policies. Furthermore, $\pi_D$ has likely identified some ``weak points" in $f_D$ and is exploiting those to induce catastrophic failure whereas $\pi_R$ does not exploit these vulnerabilities.

\subsection{Network Layer Analysis} \label{layers}

\begin{figure}[t]
\setlength\extrarowheight{25pt}
\setlength{\tabcolsep}{0pt}
\renewcommand{\arraystretch}{1.5}
\begin{center}
\begin{tabular}{>{\centering\arraybackslash}m{0.3cm}ccc}
& {\bf \Large Before} & {\bf \Large After} & {\bf \Large Single} \\
\rotatebox{90}{\parbox{2cm}{\centering \textbf{\ \ \ \ RN18}}} &  \subfig{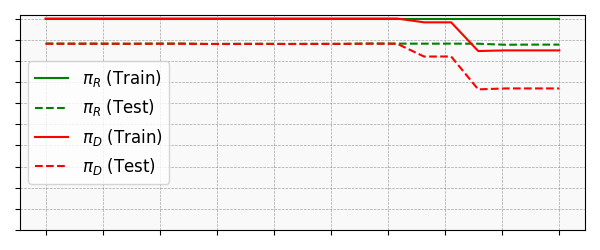} & \subfig{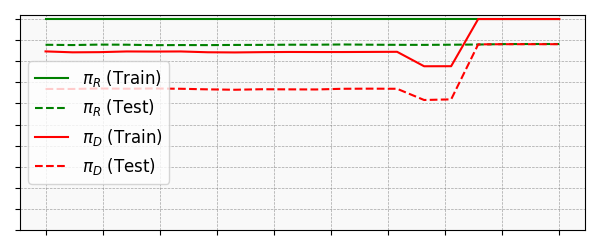} & \subfig{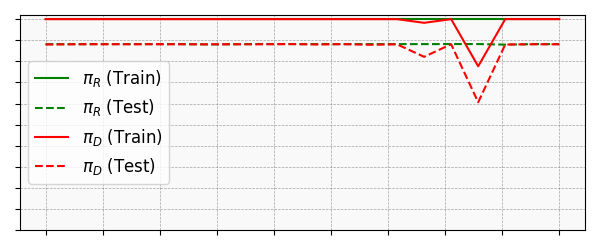}  \\ 
\rotatebox{90}{\parbox{2cm}{\centering \textbf{\ \ \ \ MNv4}}} &  \subfig{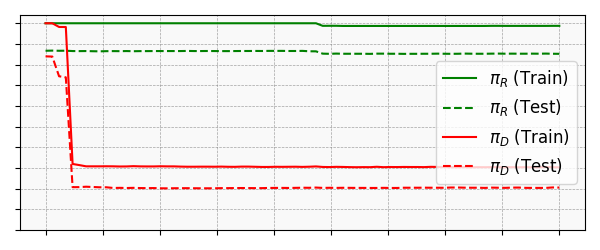} & \subfig{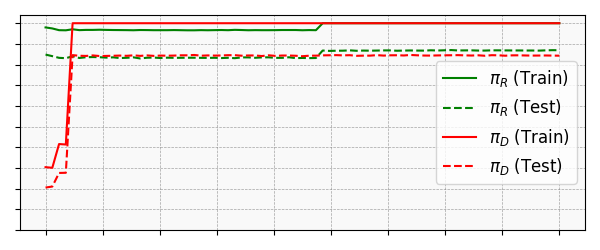} & \subfig{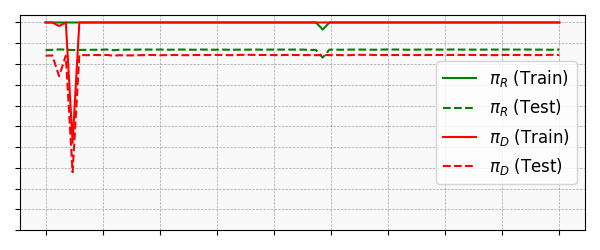} \\ 
\rotatebox{90}{\parbox{2cm}{\centering \textbf{\ \ \ \ RGX1.6}}} &  \subfig{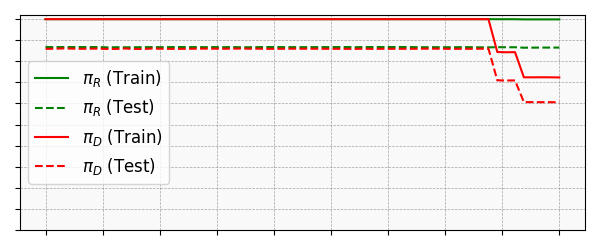} & \subfig{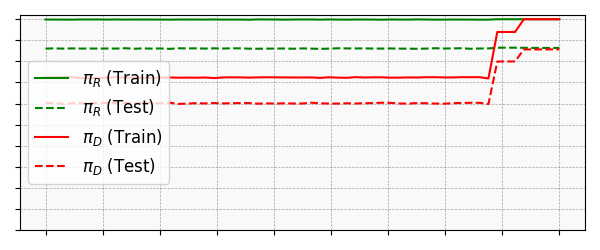} & \subfig{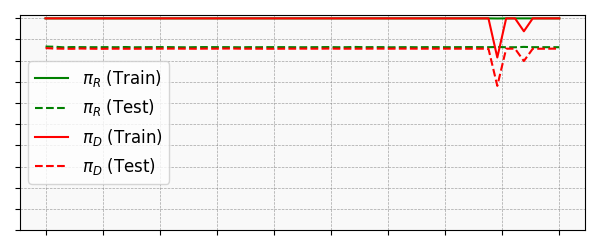} \\[-11ex]
& \multicolumn{3}{c}{Relative Layer Depth (Input Layer $\rightarrow$ Pre-Classifier Layer)}
\end{tabular}
\end{center}
\caption{Train and test accuracy of the highlight models from Table \ref{tab:main_policies} when certain layers are quantized and others are left at full-precision. ``Before" quantizes all layers \textit{before} that point (the relative layer on the x-axis), ``after" quantizes all layers \textit{after} that point, and ``single" quantizes just that single layer. All other layer activations remain at full-precision.}
\label{fig:layer_analysis}
\end{figure}

Knowing there could exist some layers whose activations are more susceptible to quantization error than others, we explored this further by examining only quantizing certain subsets of layer activations and leaving others at full-precision. We investigated three different schemes for isolated quantization: ``before", ``after", and ``single". Given some layer $l \in [1, L]$, ``before" will quantize all layers \textit{before} $l$ with everything else being full-precision. Conversely, ``after" will quantize all layers \textit{after} $l$ with the previous layers left at full-precision (\ie, the inverse of ``before"). Finally, ``single" will quantize the activations output from just that individual layer and all other layers remain at full-precision. These different variations allows us to isolate the effects of quantization on specific layers and their impact and cascading effects on the overall network. Once a network is quantized using any of the aforementioned configurations, we compute the accuracy on the train and test sets, which should allow us to see where degradation begins to occur in a model.

We plot the scores for all dynamically quantizable layer activations in the highlighted ResNet18, MobileNetV4-Conv-Medium, and RegNetX-1.6GF models and present these results in Fig.~\ref{fig:layer_analysis}. Each of the models appear to have different points at which the quantization ultimately leads to catastrophic failure. For ResNet18, this occurs about $75\%$ of the way through the network and in roughly the last $10\%$ of the RegNetX-1.6GF model. Coincidentally, this happens to be the last 6-10 layers for both networks. As for the MobileNetV4-Conv-Medium network, we interestingly observe much different behavior where the degradation stems from the earlier layers of the model. We note that the MobileNetV4 family of models is specifically designed for mobile/edge deployment compared to ResNet and RegNet which were developed for more general usage. Thus, there could be architectural differences and certain design choices that lead to this difference in behaviors. We leave further investigation into these differences as future work. Looking at the ``single" plots, we see that if we quantize just the layers that are inflection points in ``before" and ``after", these layers alone are enough to cause significant performance reduction. Thus, one could consider leaving these crucial layers at full-precision and only quantizing what remains.

\begin{figure}[t]
\setlength{\tabcolsep}{0pt}
\begin{center}
\begin{tabular}{cc}
\multirow{2}{*}{{\bf R)}}\rotatebox{90}{\parbox[t]{2.7cm}{\centering \textbf{FP}}} \ & \includegraphics[height=2.7cm,trim={0 5cm 0 0},clip]{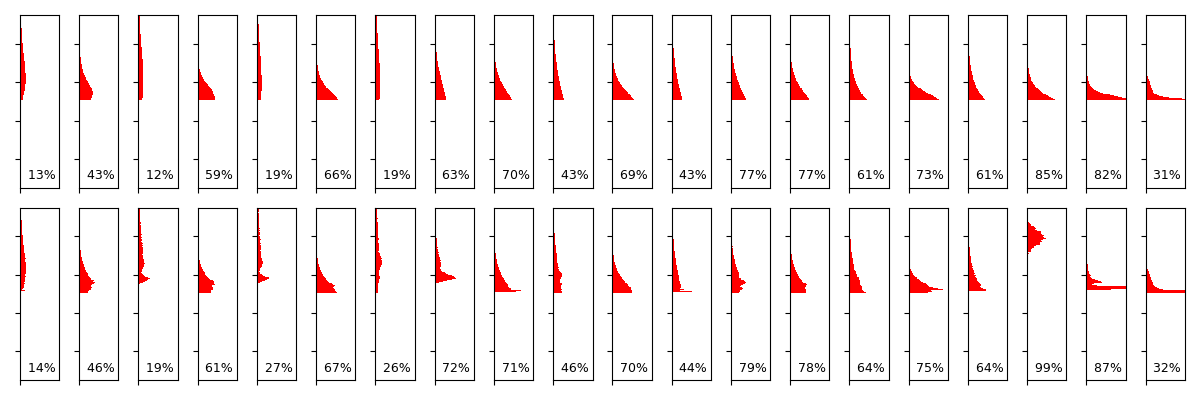}   \\
\multirow{2}{*}{\ \ \ }\rotatebox{90}{\parbox[t]{2.7cm}{\centering \textbf{Q}}} & \includegraphics[height=2.7cm,trim={0 0 0 5cm},clip]{images/histograms/resnet18_R_hist.png}   \\\hline
& \\ [-2ex]
\multirow{2}{*}{{\bf D)}}\rotatebox{90}{\parbox[t]{2.7cm}{\centering \textbf{FP}}} \  & \includegraphics[height=2.7cm,trim={0 5cm 0 0},clip]{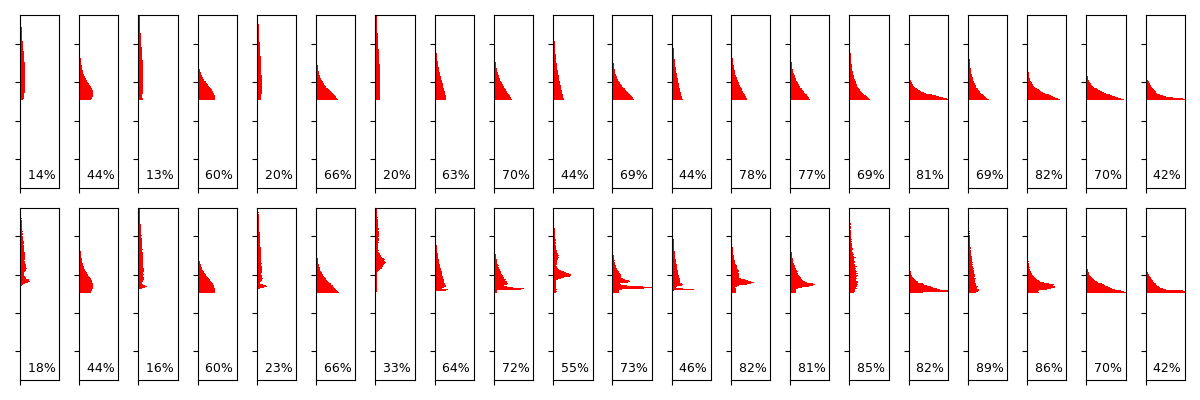}   \\
\multirow{2}{*}{\ \ \ }\rotatebox{90}{\parbox[t]{2.7cm}{\centering \textbf{Q}}} & \includegraphics[height=2.7cm,trim={0 0 0 5cm},clip]{images/histograms/resnet18_D_hist.png}   \\ [-1.6ex]
& \makebox[15.1cm][c]{%
  \ \makebox[0pt][c]{1}\hfill%
  \makebox[0pt][c]{2}\hfill%
  \makebox[0pt][c]{3}\hfill%
  \makebox[0pt][c]{4}\hfill%
  \makebox[0pt][c]{5}\hfill%
  \makebox[0pt][c]{6}\hfill%
  \makebox[0pt][c]{7}\hfill%
  \makebox[0pt][c]{8}\hfill%
  \makebox[0pt][c]{9}\hfill%
  \makebox[0pt][c]{10}\hfill%
  \makebox[0pt][c]{11}\hfill%
  \makebox[0pt][c]{12}\hfill%
  \makebox[0pt][c]{13}\hfill%
  \makebox[0pt][c]{14}\hfill%
  \makebox[0pt][c]{15}\hfill%
  \makebox[0pt][c]{16}\hfill%
  \makebox[0pt][c]{17}\hfill%
  \makebox[0pt][c]{18}\hfill%
  \makebox[0pt][c]{19}\hfill%
  \makebox[0pt][c]{20}%
} \\
& Relative Layer Depth (Input Layer $\rightarrow$ Pre-Classifier Layer)
\end{tabular}
\end{center}
\caption{Histogram of layer activation values (with the bin corresponding to $0$ and near-$0$ values removed) before and after quantization for the highlighted ResNet18 models in Table \ref{tab:main_policies}, corresponding to the FP and Q rows, respectively.}
\label{fig:feature_histograms}
\end{figure}

We take a step further into investigating models on a per-layer basis by plotting the histograms of layer activations for specifically the highlighted ResNet18 networks, which had nearly identical full-precision scores but drastically different performance when quantized. We set the bins of each histogram in the range $[-8, +8]$ with bin widths of $0.01$, thus $1600$ total bins. Furthermore, since the activations of this network are post-ReLU, these activations tend to be fairly sparse with the large majority of values being $0$. To account for this, we remove the bin corresponding to $0$ and re-normalize the histogram. We still include the percentage of features that were in this bin at the bottom of each plot which we refer to as a ``sparsity measure". Note that small values collapsing to $0$ upon quantization is a common issue \cite{Gholami2022a}, thus this is an important score to consider for our analysis as it could be a large factor in causing catastrophic failure.

As shown in Fig.~\ref{fig:feature_histograms}, the histograms for $\{f_D, \pi_D\}$ do tend to change more frequently and intensely compared to $\{f_R, \pi_R\}$. There is still one layer (18) in $\{f_R, \pi_R\}$ that encounters a large increase in $0$ or near-$0$ values after quantization, but the model has learned some way to lessen the effect of this layer. Conversely, for $\{f_D, \pi_D\}$, many of the layers incur dramatic changes in their histograms from quantization, especially in layers 15 and 17 which gain $15-20\%$ in their sparsity measures. These layers also happen to align with the inflection points in the previous experiment, which could suggest that having long-tailed activations increases the potential for critical failure. In other words, having a large majority of activations near $0$ and few strong outliers could enable a significant portion of values to collapse to $0$, depending on the quantization scheme. In this scenario, using a quantization scale equal to the $n$-th quantile, rather than the min-max or abs-max, may be more favorable.

\subsection{Generalization \& Robustness Analysis} \label{robustness}

\begin{figure}[t]
    \begin{center}
    \includegraphics[width=16cm]{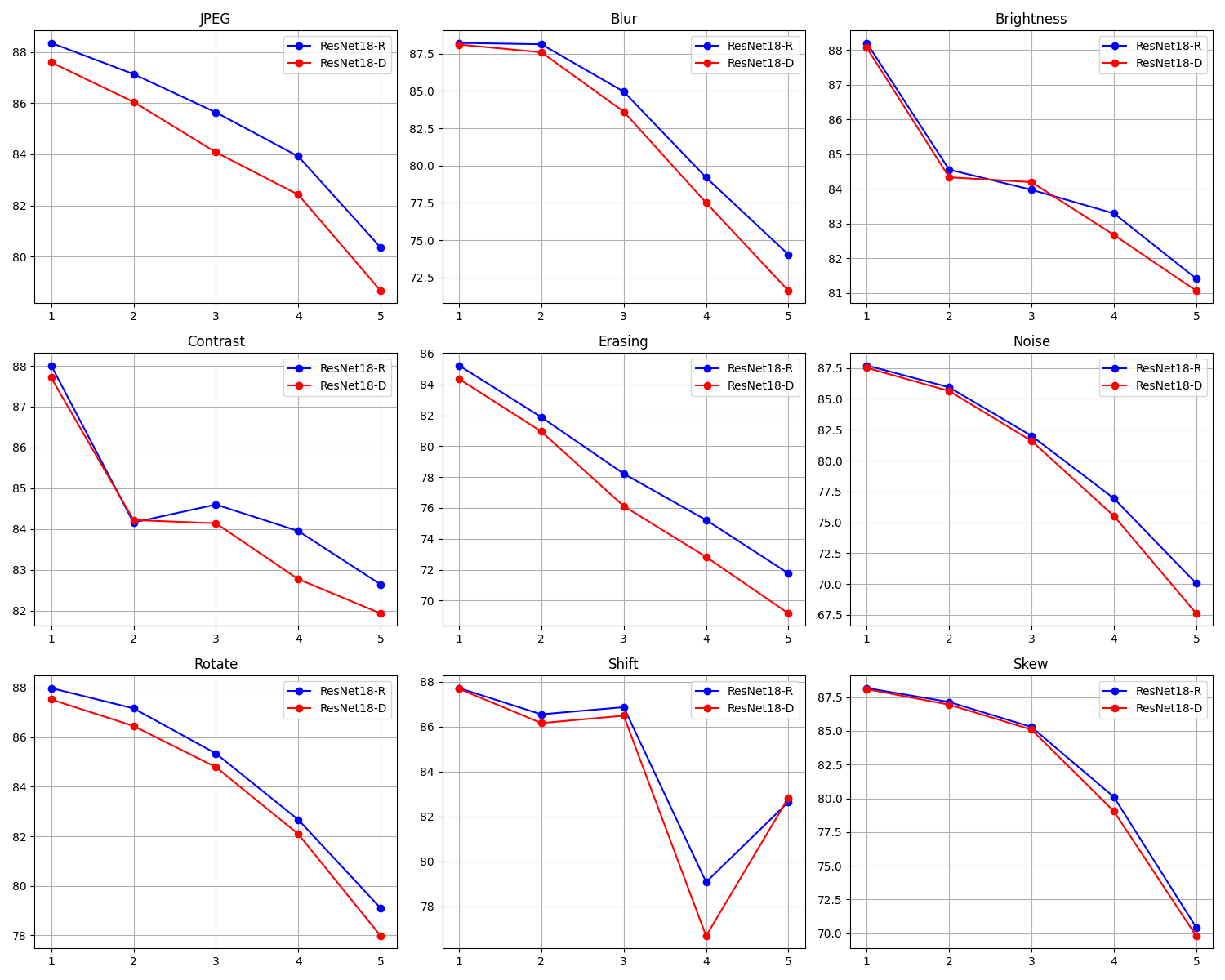}
    \end{center}
    \caption{A generalization/robustness analysis for the highlighted ResNet18 models from Table \ref{tab:main_policies} when transforming/corrupting the test set images with common operations. Degree and test accuracy are the $x$ and $y$ axes, respectively.}
    \label{fig:transforms}
\end{figure}

To save time, money, and other resources, it would be beneficial to identify models that are susceptible to DPTQ-based catastrophic failure before they are deployed on a safety-critical platform. Thus, we question whether there are any potential indicators of brittleness in the full-precision models. To explore this we performed an initial study by examining the generalization/robustness capabilities of the highlighted ResNet18 models to various common input transformations/corruptions. Such alterations include: JPEG compression, Gaussian blur, brightness jittering, contrast jittering, random block erasing, Gaussian noise, rotation, skewing, and distribution shift. Random erasing refers to the common random erasing data augmentation \cite{Zhong2020a} which selects random rectangular regions in an image and sets the values in that area to Gaussian noise. Distribution shift can take on many different interpretations in practice. Here, we simulate an input distribution shift by perturbing the statistics used for normalizing an input image before it is passed through the network. Moreover, for each transformation/corruption, we include a ``degree" parameter which increases the effect of the operation as the degree increases. Examples include decreasing the quality for JPEG compression, increasing the standard deviation for Gaussian blur and noise, larger rectangular regions for random erasing, etc. Thus, as the degree gets larger, we should expect a decrease in accuracy as the images are becoming more corrupted.

We see in Fig.~\ref{fig:transforms} that the robust network $f_R$ does tend to perform better than the brittle network $f_D$ for many of the transformation/corruptions. Note that the models performed similarly for the unperturbed test set ($88.17\%$ vs. $88.09\%$ for $f_R$ and $f_D$, respectively). Although the differences between $f_R$ and $f_D$ are relatively small for all transformations, there is still a noticeable trend, which would become clearer (\ie, more accurate) with more trials via random seeds. Our results are encouraging that potentially brittle networks could be identified to some capacity and hopefully motivate future work in the area.

\section{Conclusion} \label{sec:conclusion}

We identified and investigated the existence of catastrophic failure stemming from dynamic post-training quantization, which is characterized by drastic performance reduction when the desired quantization approach is applied, to the extent that the model (or overall system) is rendered incompetent. Through our experimentation and analyses, we conducted an initial exploration into the potential for and extent of such a situation. More specifically, our results demonstrated the ability to obtain two models of identical network architecture that achieved similar performance levels at full-precision, but exhibited up to a $25\%$ difference in their scores when quantized. Furthermore, we performed several studies investigating the impact of individual layer activations to the overall degradation and presented a preliminary appraisal of full-precision models to assess the potential for indicators that a model could ultimately fail once quantized. This work presents an inaugural investigation into potential harm stemming from dynamic post-training quantization, with several avenues for future work and examination. Overall, we find this to be an immensely important topic that demands careful consideration when contemplating the use of any quantization scheme in a safety-critical environment.

\bibliography{main} 
\bibliographystyle{spiebib} 

\end{document}